\documentclass[journal,article,accept,pdftex,moreauthors]{Definitions/mdpi}

\usepackage{algorithmic}
\usepackage{amsmath}
\usepackage{amssymb}
\usepackage{graphicx}
\usepackage{booktabs}

\usepackage{algorithm}
\usepackage{geometry}
\usepackage{listings}

\usepackage{svg}
\usepackage{subcaption}
\usepackage{float}
\setcitestyle{square, numbers} 

\firstpage{1} 
\pubvolume{1}
\issuenum{1}
\articlenumber{0}
\pubyear{2026}
\copyrightyear{2025}

\hreflink{https://doi.org/} 

\pdfoutput=1 

\Title{Physics-Informed Diffusion Model for Generating Synthetic Extreme Rare Weather Events Data}

\Author{Marawan Yakout $^{1}$\orcidA{}, Tannistha Maiti $^{2}$ , Monira Majhabeen $^{2}$ and Tarry Singh $^{2}$ }

\AuthorNames{Marawan Yakout, Tannistha Maiti, Monira Majhabeen and Tarry Singh}

\address{
$^{1}$ \quad Department of Computer Science, University of London, London;\\
$^{2}$ \quad Deepkapha AI Lab, Assen, The Netherlands;\\
}

\corres{Correspondence: mmyay1@student.london.ac.uk; yakout@marawan.net (M.Y.); tannistha.maiti@deepkapha.com (T.M.)}

\abstract{Data scarcity is a primary obstacle in developing robust Machine Learning (ML) models for detecting rapidly intensifying tropical cyclones. Traditional data augmentation techniques (rotation, flipping, brightness adjustment) fail to preserve the physical consistency and high-intensity gradients characteristic of rare Category 4-equivalent events, which constitute only 0.14\% of our dataset (202 of 140,514 samples). We propose a physics-informed diffusion model based on the Context-UNet architecture to generate synthetic, multi-spectral satellite imagery of extreme weather events. Our model is conditioned on critical atmospheric parameters such as average wind speed, type of Ocean and stage of development (early, mature, late etc)---the known drivers of rapid intensification. Using a controlled pre-generated noise sampling strategy and mixed-precision training, we generated $16\times16$ wind-field samples that are cropped from multi-spectral satellite imagery which preserve realistic spatial autocorrelation and physical consistency. Results demonstrate that our model successfully learns discriminative features across ten distinct context classes, effectively mitigating the data bottleneck. Specifically, we address the extreme class imbalance in our dataset, where Class 4 (Ocean 2, early stage with average wind speed 50kn hurricane) contains only 202 samples compared to 79,768 samples in Class 0. This generative framework provides a scalable solution for augmenting training datasets for operational weather detection algorithms. The average  Results yield an average Log-Spectral Distance (LSD) of 4.5dB, demonstrating a scalable framework for enhancing operational weather detection algorithms.}


\begin{document}





\section{Introduction}


Data scarcity poses a significant challenge for extreme weather events in the context of developing Machine Learning (ML) models for rare weather events. Forecasting of physical systems over both space and time is a major problem that has many real-world applications, including natural sciences, transportation, and energy systems. Numerical Weather Prediction (NWP) systems currently predict the weather using complex physical models and large supercomputers \cite{bauer_quiet_2015}. In the past decade, the data generated from spatiotemporal Earth Observations have increased in an unprecedented way, allowing data-driven forecasting models to increase and improve using deep learning techniques.  \newline

Currently, ML models' data enhancement is used for the detection of rapidly intensifying tropical cyclones. However, these models are not able to capture extreme events due to a lack of data, generating physically implausible predictions. This is mainly due to the rarity of such events, blurred forecasts, and the need for massive data to train such models. In addition to the high computational cost of training such models \cite{gao2023prediff, bauer_quiet_2015}. \\ 

However, the most extreme events (e.g, \textbf{Category 5} hurricanes that undergoing rapid intensification just before landfall) are statistically rare.\\ Recently, Machine Learning Weather Prediction (MLWP) models have emerged, challenging the performance of the existing NWP systems' approaches. These models are not physics-based but mainly data-driven, mainly made as a result of deep learning algorithmic advancements and discoveries. \cite{andrae_continuous_2024} \\

We propose leveraging diffusion models to generate synthetic data samples of rare weather events to augment training data for other AI/ML models. A critical requirement is ensuring these synthetic images exhibit physical consistency, real enough with the known physics of the atmosphere, to ensure that the synthetic data do not confuse downstream model training.\\

Moreover, the specific atmospheric conditions that lead to the Rapid Intensification (RI) of tropical cyclones are well known. Our generative model is conditioned on parameters defining RI. (e.g., low vertical wind shear + high ocean heat content). We leverage multimodal satellite data from NASA Global Precipitation Measurement (GPM) satellite data, combined with NASA Geostationary Operational Environmental Satellite (GOES) imagery RI data to identify intense convective bursts \cite{GPMData}.\\

This research designs a physics-informed diffusion model to generate synthetic, multispectral satellite images of rare tropical cyclones undergoing rapid intensification, conditioned on known atmospheric parameters, to augment robust training of operational detection algorithms. 


\section{Related Work}

Our work intersects several research areas: tropical cyclone prediction, data augmentation for imbalanced datasets, generative modeling, and physics-informed machine learning. We review relevant work in each area and position our contributions.

\subsection{Tropical Cyclone Rapid Intensification}
Rapid intensification (RI) of tropical cyclones---defined as wind speed of at least 30 knots (15 m/s) in 24 hours \cite{kaplan2003large} ---remains one of the most challenging problems in operational meteorology. The atmospheric and oceanic conditions favorable for RI are well-established: low vertical wind shear, high sea surface temperatures (>26$^\circ$C), high ocean heat content, moist mid-level atmosphere and favorable upper-level outflow \cite{kaplan2003large}. Despite this physical understanding, accurately predicting which storms will undergo RI remains difficult due to the complex interplay of environmental factors and internal storm dynamics.\\

Traditional statistical approaches such as the Statistical Hurricane Intensity Prediction Scheme (SHIPS) \cite{demaria2005further} incorporate environmental predictors but show limited skill for RI forecasting. Recent deep learning approaches have shown promise: Wimmers et al. \cite{wimmers2019deep} use deep learning with passive microwave satellite imagery to estimate TC intensity, while Lee et al. \cite{lee_tropical_2019} apply multi-dimensional CNNs to geostationary satellite data. Chen et al. \cite{chen2019machine} provide a comprehensive review of machine learning approaches in tropical cyclone forecast modeling. However, these methods typically focus on current intensity estimation rather than RI prediction, and their performance degrades for extreme intensification events.

A fundamental limitation of data-driven approaches is the scarcity of RI events in historical records. The problem is particularly acute for Category 5-equivalent tropical cyclones, which constitute less than 0.2\% of observations. In our dataset of 140,514 storm observations, the most extreme class (Class 4) contains only 202 samples compared to 79,768 in the baseline class---a nearly 400-fold imbalance. This severe class imbalance prevents supervised learning models from effectively capturing the subtle signatures preceding the most dangerous storms---precisely the problem our work addresses through synthetic data generation.

\subsection{Class Imbalance and Data Augmentation}
Class imbalance poses fundamental challenges for supervised learning, particularly when minority classes represent the most consequential events. For image data, augmentation typically involves geometric transformations (rotation, flipping, cropping) and photometric adjustments (brightness, contrast, color jittering) \cite{shorten2019survey}. While effective for general computer vision tasks, these techniques have significant limitations for atmospheric data.

Arbitrary rotation of hurricane imagery violates the relationship between latitude and Coriolis-induced rotation direction (counterclockwise in the Northern Hemisphere, clockwise in the Southern Hemisphere). Photometric transformations corrupt the physical relationship between pixel intensity and meteorological quantities like wind speed or precipitation rate measured by satellite sensors \cite{GPMData,GOESData}. More fundamentally, traditional augmentation creates variations of existing samples rather than expanding coverage of the underlying data manifold. For extreme events represented by only hundreds of examples among hundreds of thousands, this limitation is particularly severe---augmented samples remain tethered to the specific storms in the training set rather than exploring the broader space of physically plausible extreme events.

\subsection{Generative Models for Data Augmentation}

Generative Adversarial Networks (GANs) \cite{goodfellow_generative_2014} learn to generate synthetic samples by training a generator network to fool a discriminator network. Conditional GANs \cite{mirza2014conditional} extend this framework to allow controlled generation based on class labels or other conditioning information. While GANs have achieved impressive results in high-fidelity image synthesis, they suffer from well-known training instabilities, mode collapse (failure to capture full data diversity), and difficulty generating samples for underrepresented classes---problems that are exacerbated when minority classes are already severely underrepresented.

Denoising Diffusion Probabilistic Models (DDPM) \cite{ho_denoising_2020} offer an alternative generative framework with more stable training dynamics. By gradually adding Gaussian noise to data over many timesteps and learning to reverse this process, diffusion models achieve high sample quality and diversity. Recent work has demonstrated that diffusion models can surpass GANs in image synthesis quality and sample diversity \cite{dhariwal_diffusion_2021}. Latent diffusion models \cite{rombach_high-resolution_2022} extend this approach to high-resolution generation by operating in compressed latent spaces, improving computational efficiency.

\subsection{Application to weather and climate data} 
Recent work has begun exploring diffusion models for meteorological applications. GenCast \cite{price_gencast_2024} applies diffusion models to ensemble weather forecasting, demonstrating competitive skill with traditional numerical weather prediction. Leinonen et al. \cite{leinonen2023latent} use latent diffusion models for precipitation nowcasting with accurate uncertainty quantification. \newpage

Ravuri et al. \cite{ravuri2021skilful} demonstrate that deep generative models can produce skillful precipitation forecasts. These works validate the feasibility of diffusion models for generating physically plausible atmospheric fields, though they focus on forecasting rather than synthetic data augmentation for class imbalance---the gap our work addresses.The stable training properties and sample diversity of diffusion models make them particularly attractive for augmenting minority classes. However, their application to scientific data requires careful consideration of domain-specific constraints to ensure generated samples are not only statistically plausible but also physically realistic. 

\subsection{Physics-informed Generation}
Incorporating physical knowledge into machine learning models has gained increasing attention across scientific and engineering domains. For atmospheric modeling, physical consistency is particularly important. Wind fields should satisfy mass continuity, respect geostrophic balance at large scales, and maintain realistic spatial autocorrelation structures.\\

Our work makes several novel contributions relative to existing literature:
\begin{quote}
\begin{quote}

\begin{itemize}
    \item \textbf{Diffusion for extreme weather enhancement:} Building on recent 
    applications of diffusion models to weather data 
    \cite{price_gencast_2024,leinonen2023latent,ravuri2021skilful}, we present 
    the first work that specifically targets synthetic data enhancement for 
    extreme weather class imbalance rather than forecasting. While prior work 
    validates that diffusion models can generate physically plausible atmospheric 
    fields, we extend this to addressing the severe data scarcity problem in 
    extreme event prediction.

    \item \textbf{Addressing severe class imbalance:} We demonstrate effective 
    generation for a minority class with only 202 samples within a dataset of 
    140,514 observations---a nearly 400-fold imbalance more severe than typically 
    addressed in the machine learning literature. Our pre-generated noise 
    strategy ensures fair representation of rare classes during training, 
    addressing a subtle but important challenge when applying diffusion models 
    to severely imbalanced datasets.
    
    \item \textbf{Comparison to GAN-based approaches.} As discussed in Section 2.3, compared to standard GAN-based augmentation~\cite{goodfellow_generative_2014}, our physics-informed diffusion approach avoids mode collapse and yields greater sample diversity---properties that are crucial for training downstream detection models that must generalize across varying storm morphologies.
\end{itemize}
\end{quote}
\end{quote}


\section{Proposed Workflow Overview}

As shown in the figure \ref{fig:flowchart} below, our approach consists of three main stages. In the forward process (left), clean wind field data $x_0$ ---($16 \times 16$ spatial resolution) progressively corrupted by adding Gaussian noise following the schedule $q(x_t|x_0)$, utilizing a pre-generated noise strategy where $\epsilon \sim \mathcal{N}(0, I)$ is stored and consistently reused across training epochs to ensure fair representation of rare Class 4 samples. The training phase (center) employs a Context-UNet architecture that takes three inputs: the noisy sample $x_t$, a sinusoidal timestep embedding t, and a physics-aware context vector c encoding atmospheric parameters (wind shear, ocean heat content, SST anomalies) as one-hot classes. The U-Net (with 64 base features) predicts the noise component $\epsilon_\theta(x_t, t, c)$, which is optimized via MSE loss to match the actual noise.\newpage

The reverse process (right) uses the trained model to iteratively denoise pure Gaussian noise $x_T$ over 500 steps following $p_\theta(x_{t-1}|x_t)$, conditioned on the desired physics parameters, ultimately generating novel synthetic extreme weather events $\hat{x}_0$ that maintain physical consistency while addressing the severe class imbalance in the original dataset.

\begin{figure}[H]
    \centering
    \includegraphics[width=1\linewidth]{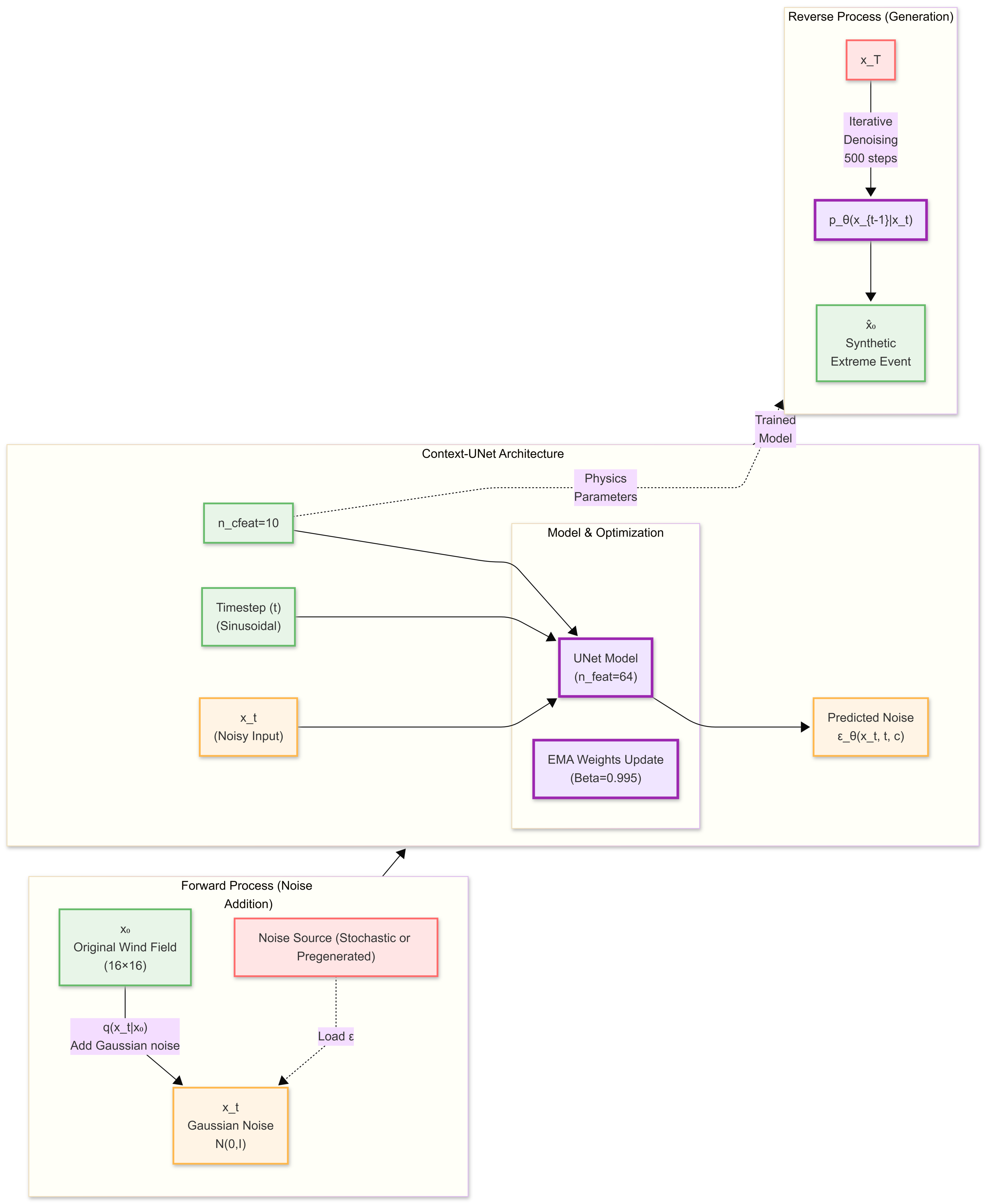}
    \caption{Figure illustrates the three-phase pipeline for generating synthetic extreme events: Forward Process: $x_0$ (original wind field) is transformed into Gaussian noise $x_t$ using a pre-generated noise strategy. Context-UNet Training: A UNet ($n\_feat=64$) predicts noise by conditioning the noisy input ($x_t$) on sinusoidal time steps ($t$) and physics-based context ($c$), such as one-hot encoded wind shear classes. Reverse Process: The trained model performs 500 iterative denoising steps starting from $x_T$ to produce a synthetic extreme event $x^{\hat{}}_0$.}
    \label{fig:flowchart}
\end{figure}

\section{Model Architecture}

The implementation utilizes the Context-UNet architecture, which is trained on single-channel $16 \times 16$ spatial wind data with pre-generated noise samples.  This framework is  scalable to higher spatial resolutions, though at increased computational and memory cost. In the following,  we detail all hyperparameters, architectural choices, optimization strategies, and the mathematical framework underlying the forward and reverse diffusion processes.

\subsection{Context U-Net Configuration}

The fundamental model used is Context U-Net  \cite{Ronneberger2015}, adapted for diffusion modeling with the following specifications:

\begin{quote}
    \begin{itemize}
        \item \textbf{Input channels}: \(c_{\text{in}} = 1\) (single-channel grayscale wind data)
        \item \textbf{Base feature channels}: \(n_{\text{feat}} = 64\)
        \item \textbf{Context feature channels}: \(n_{\text{cfeat}} = 10\) (conditional generation capacity)
        \item \textbf{Spatial resolution}: \(h = w = 16\) ($16 \times 16$ pixel grid)
    \end{itemize}
\end{quote}
 
The U-Net architecture comprises  an encoder (contracting path) and a decoder (expanding path), with skip connections that preserve spatial information across hierarchical feature representations.   The model predicts the noise vector \(\boldsymbol{\epsilon}_\theta(\mathbf{x}_t, t)\) conditioned on the noisy input  \(\mathbf{x}_t\) and the normalized timestep \(t/T\).

\subsection{Timestep Conditioning}

The discrete timestep \(t \in \{1, 2, \ldots, T\}\) is normalized to \([0, 1]\) and embedded in the network through sinusoidal positional encodings inspired by Transformer \cite{vaswani2017attention} architectures. \\

For a timestep \(t\) and embedding dimension index \(i\), the positional encoding can be presented as the following:

\begin{equation}
\text{PE}(t, 2i) = \sin\left(\frac{t}{10000^{2i/d}}\right), \quad \text{PE}(t, 2i+1) = \cos\left(\frac{t}{10000^{2i/d}}\right)
\end{equation}
where \(d\) is the embedding dimensionality. \\

The normalized timestep $\hat{t} = t/T \in (0, 1]$ is converted into a high-dimensional 
sinusoidal embedding $\text{PE}(\hat{t}) \in \mathbb{R}^d$ where each dimension alternates 
between sine and cosine functions with exponentially increasing frequencies. Specifically, 
the embedding at dimension index $i$ is computed as $\sin(\hat{t}/10000^{2i/d})$ for even 
dimensions and $\cos(\hat{t}/10000^{2i/d})$ for odd numbered dimensions. \\

\begin{quote}
\begin{quote}
\begin{itemize}
    \item$\hat{t}$ $\rightarrow$ Normalized timestep $(0, 1]$.
    \item$i$ $\rightarrow$ Dimension index of the embedding, with range $0$ to $d/2 - 1$.
    \item$2i$ $\rightarrow$ Even dimension index.
    \item$2i+1$ $\rightarrow$ Odd dimension index.
    \item$d$ $\rightarrow$ Total embedding dimensionality (e.g., 256, 512)
    \item$10000^{2i/d}$ $\rightarrow$ Frequency scaling term.
\end{itemize}
\end{quote}
\end{quote}

The resulting embedding is processed through a learned multi-layer perceptron and integrated into the U-Net's convolutional blocks enabling the model to condition its denoising operations on the current noise level. \\

\section{Diffusion Process Formulation}

\subsection{Pre-generated Noise Strategy}

Unlike standard Denoising Diffusion Probabilistic Models (DDPM) \cite{ho_denoising_2020}that sample noise dynamically during each training iteration, this implementation utilizes a pre-generated noise strategy. Noise samples $\epsilon \sim \mathcal{N}(0, \mathbf{I})$ are generated offline and stored in a single large-scale binary file to ensure reproducibility across training runs and maintain pairing integrity with the sparse dataset. \\

As illustrated in Figure \ref{fig:label_dist}, the extreme scarcity of certain event classes---specifically Class 4 with only 202 samples compared to 79,768 in Class 0. This creates a nearly 400-fold imbalance and requires a highly controlled training environment. By pre-assigning a dedicated noise sequence to each image, we ensure that the model's exposure to rare intensities is consistent across epochs.\\

The noise tensor is structured with the shape $(N_{\text{images}}, T, C, H, W)$, where $N_{\text{images}} = 140,514$ is the total dataset size, $T=500$ represents the diffusion timesteps, $C=1$ for grayscale atmospheric data, and the spatial dimensions are $16 \times 16$.\\

\begin{figure}[h]
    \centering
    \includegraphics[width=\textwidth]{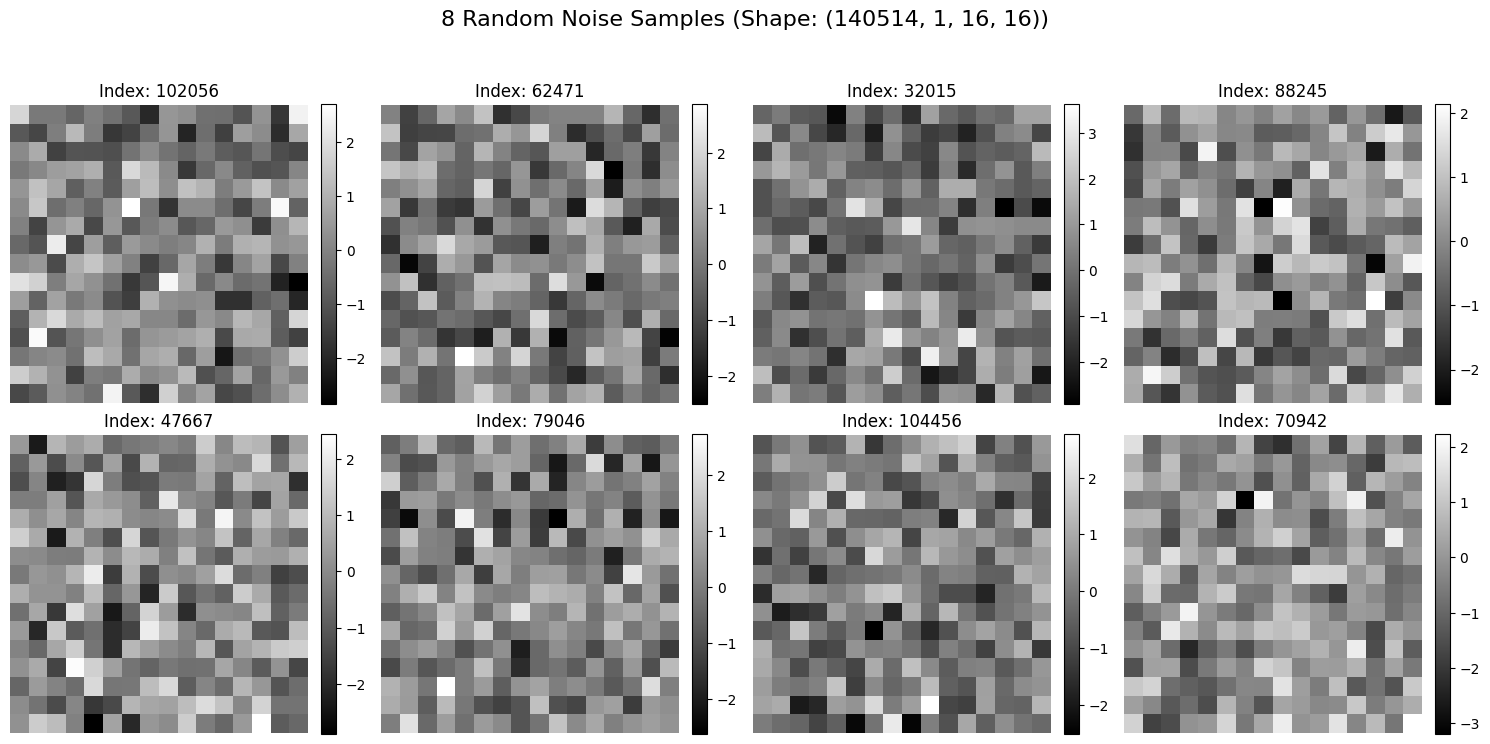} 
    \caption{Visual and statistical verification of eight random noise samples. Each patch represents a $16 \times 16$ realization of $\epsilon \sim \mathcal{N}(0, 1)$. The consistent Gaussian distribution across samples ensures that rare event detection is not biased by noise artifacts.}
    \label{fig:noise_samples}
 \end{figure}

To verify the integrity of the generation, eight random samples were extracted and analyzed. As shown in Figure \ref{fig:noise_samples}, each $16 \times 16$ patch exhibits the expected "white noise" characteristics of a standard normal distribution. \\

\subsection{Forward Diffusion Process}

The forward process gradually corrupts clean data $\mathbf{x}_0 \sim q(\mathbf{x}_0)$ by adding Gaussian noise over $T$ discrete timesteps following a Markov chain:

\begin{equation}
q(\mathbf{x}_t | \mathbf{x}_{t-1}) = \mathcal{N}(\mathbf{x}_t; \sqrt{1 - \beta_t} \mathbf{x}_{t-1}, \beta_t \mathbf{I})
\end{equation}

where $\beta_t$ is the variance schedule controlling noise addition at each step. However, the code implementation employs an optimization: instead of iterating sequentially as $\mathbf{x}_0 \to \mathbf{x}_1 \to \mathbf{x}_2 \to \cdots \to \mathbf{x}_t$, it computes $\mathbf{x}_t$ directly in a single step via the reparameterization trick, significantly reducing computational overhead.

where $\epsilon \sim \mathcal{N}(0, I)$ is Gaussian noise.

\subsection{Reverse Diffusion}

The reverse diffusion process is then used, the model learns to denoise samples by iteratively removing noise, reconstructing \(\mathbf{x}_0\) from pure Gaussian noise \(\mathbf{x}_T \sim \mathcal{N}(\mathbf{0}, \mathbf{I})\). The reverse transition is modeled as a Gaussian distribution with learned parameters:

The reverse process learns to denoise samples by predicting the noise component at each timestep:

\begin{equation}
p_\theta(x_{t-1} | x_t) = \mathcal{N}(x_{t-1}; \mu_\theta(x_t, t), \Sigma_\theta(x_t, t))
\end{equation}

The mean is computed as:

\begin{equation}
\mu_\theta(x_t, t) = \frac{1}{\sqrt{\alpha_t}}\left(x_t - \frac{1-\alpha_t}{\sqrt{1-\bar{\alpha}_t}}\epsilon_\theta(x_t, t, c)\right)
\end{equation}

where $\epsilon_\theta$ is the neural network that predicts the noise, and $c$ represents optional context conditioning.

\subsubsection{Linear Variance Schedule}

We employ a linear variance schedule with the following parameters:
\begin{quote}
    \begin{quote}
        \begin{itemize}
            \item \textbf{Timesteps}: \(T = 500\)
            \item \textbf{Initial variance}: \(\beta_1 = 1 \times 10^{-3} = 0.001\)
            \item \textbf{Final variance}: \(\beta_T = 2 \times 10^{-2} = 0.02\)
        \end{itemize}
    \end{quote}
\end{quote}

The schedule is then constructed via linear interpolation:

\begin{equation}
\beta_t = \beta_1 + \frac{\beta_T - \beta_1}{T} \cdot (t - 1), \quad t \in \{1, 2, \ldots, T\}
\end{equation}

This creates a sequence \(\{\beta_t\}_{t=1}^{T}\) that linearly increases from \(\beta_1\) to \(\beta_T\).

\subsubsection{Derived Coefficients}

The complementary variance (signal retention coefficient) is defined as:

\begin{equation}
\alpha_t = 1 - \beta_t
\end{equation}

The cumulative product of $\alpha_t$ values, denoted $\bar{\alpha}_t$, is computed as:

\begin{equation}
\bar{\alpha}_t = \prod_{s=1}^{t} \alpha_s = \exp\left(\sum_{s=1}^{t} \log \alpha_s\right)
\end{equation}

with the initial condition $\bar{\alpha}_0 = 1$. This coefficient governs the signal-to-noise ratio decay throughout the diffusion trajectory. The log-sum-exp formulation is paramount for numerical stability, preventing \texttt{float32} underflow when computing products over large numbers of timesteps.

\subsubsection{Efficient Forward Perturbation}

An essential advantage of DDPM is the ability to sample \(\mathbf{x}_t\) at any arbitrary timestep \(t\) directly from \(\mathbf{x}_0\) without iterating through intermediate steps. The closed-form reparameterization can be presented as follows:

\begin{equation}
\mathbf{x}_t = \sqrt{\bar{\alpha}_t} \mathbf{x}_0 + \sqrt{1 - \bar{\alpha}_t} \boldsymbol{\epsilon}, \quad \boldsymbol{\epsilon} \sim \mathcal{N}(\mathbf{0}, \mathbf{I})
\label{eq:forward_perturbation}
\end{equation}

This is implemented in the \texttt{perturb\_input} function:

\begin{equation}
\texttt{perturb\_input}(\mathbf{x}, t, \boldsymbol{\epsilon}) = \sqrt{\bar{\alpha}_t} \mathbf{x} + \sqrt{1 - \bar{\alpha}_t} \boldsymbol{\epsilon}
\end{equation}

The noise vector \(\boldsymbol{\epsilon}\) is sourced from pregenerated noise samples.
A critical design choice in this implementation is the use of \textbf{pregenerated noise} stored in the \texttt{./pregenerated\_noise} directory. This approach differs from standard DDPM implementations where noise is sampled randomly during each training iteration.

In contrast, standard implementations generate noise dynamically:
\begin{quote}
\begin{quote}
\begin{itemize}
    \item Fresh noise samples drawn from \(\mathcal{N}(\mathbf{0}, \mathbf{I})\) each iteration
    \item Implemented as \texttt{torch.randn\_like(x)}
    \item No storage overhead, fully stochastic training
\end{itemize}
\end{quote}
\end{quote}


\subsection{Reverse Diffusion Process}

The reverse diffusion process is then used, the model learns to denoise samples by iteratively removing noise, reconstructing \(\mathbf{x}_0\) from pure Gaussian noise \(\mathbf{x}_T \sim \mathcal{N}(\mathbf{0}, \mathbf{I})\). The reverse transition is modeled as:

\begin{equation}
p_\theta(\mathbf{x}_{t-1} | \mathbf{x}_t) = \mathcal{N}(\mathbf{x}_{t-1}; \boldsymbol{\mu}_\theta(\mathbf{x}_t, t), \tilde{\beta}_t \mathbf{I})
\end{equation}

\subsubsection{Posterior Mean Estimation}

The posterior mean \(\boldsymbol{\mu}_\theta\) is derived from Tweedie's formula and the predicted noise \(\boldsymbol{\epsilon}_\theta(\mathbf{x}_t, t)\):

\begin{equation}
\boldsymbol{\mu}_\theta(\mathbf{x}_t, t) = \frac{1}{\sqrt{\alpha_t}} \left( \mathbf{x}_t - \frac{\beta_t}{\sqrt{1 - \bar{\alpha}_t}} \boldsymbol{\epsilon}_\theta(\mathbf{x}_t, t) \right)
\label{eq:posterior_mean}
\end{equation}

This equation represents the expected value of \(\mathbf{x}_{t-1}\) given \(\mathbf{x}_t\) and the model's noise prediction.

\subsubsection{Denoising with Stochasticity}

The sampling step combines the posterior mean with injected noise (except at the final timestep):

\begin{equation}
\mathbf{x}_{t-1} = \boldsymbol{\mu}_\theta(\mathbf{x}_t, t) + \sqrt{\beta_t} \mathbf{z}, \quad \mathbf{z} \sim \mathcal{N}(\mathbf{0}, \mathbf{I})
\end{equation}

For \(t = 1\) (final denoising step), we set \(\mathbf{z} = \mathbf{0}\) to obtain a deterministic reconstruction.

The implementation in \texttt{denoise\_add\_noise} follows:

\begin{equation}
\texttt{denoise\_add\_noise}(\mathbf{x}_t, t, \boldsymbol{\epsilon}_\theta, \mathbf{z}) = \frac{1}{\sqrt{\alpha_t}} \left( \mathbf{x}_t - \frac{(1 - \alpha_t)}{\sqrt{1 - \bar{\alpha}_t}} \boldsymbol{\epsilon}_\theta \right) + \sqrt{\beta_t} \mathbf{z}
\end{equation}


\section{Training Configuration}

\subsection{Optimization Hyperparameters}

\begin{table}[H]
    \centering
    \caption{Core Training Hyperparameters}
    \label{tab:hyperparams}
    \begin{tabular}{@{}ll@{}}
        \toprule
        \textbf{Parameter} & \textbf{Value} \\ 
        \midrule
        Batch size & 64 \\
        Training epochs & 120 \\
        Initial learning rate ($\eta_{\text{max}}$) & $1 \times 10^{-4}$ \\
        Minimum learning rate ($\eta_{\text{min}}$) & $1 \times 10^{-6}$ \\
        Optimizer & Adam \\
        EMA decay coefficient ($\beta_{\text{EMA}}$) & 0.995 \\
        \bottomrule
    \end{tabular}
\end{table}

\subsection{Adam Optimizer}

The Adam (Adaptive Moment Estimation) optimizer is employed with default PyTorch parameters:
\begin{quote}
\begin{quote}
\begin{itemize}
    \item \textbf{Learning rate}: \(\alpha = 1 \times 10^{-4}\)
    \item \textbf{Beta coefficients}: \(\beta_1 = 0.9\) (first moment), \(\beta_2 = 0.999\) (second moment)
    \item \textbf{Epsilon}: \(\epsilon = 1 \times 10^{-8}\) (numerical stability)
\end{itemize}
\end{quote}
\end{quote}
Adam computes adaptive learning rates for each parameter by maintaining exponential moving averages of gradients (\(m_t\)) and squared gradients (\(v_t\)):

\begin{align}
m_t &= \beta_1 m_{t-1} + (1 - \beta_1) g_t \\
v_t &= \beta_2 v_{t-1} + (1 - \beta_2) g_t^2 \\
\theta_{t+1} &= \theta_t - \alpha \frac{\hat{m}_t}{\sqrt{\hat{v}_t} + \epsilon}
\end{align}

where \(\hat{m}_t\) and \(\hat{v}_t\) are bias-corrected estimates.

\subsection{Cosine Annealing Learning Rate Schedule}

To improve convergence and generalization, we employ Cosine Annealing with warm restarts (specifically, \texttt{CosineAnnealingLR} in PyTorch):

\begin{equation}
\eta_t = \eta_{\text{min}} + \frac{1}{2} (\eta_{\text{max}} - \eta_{\text{min}}) \left(1 + \cos\left(\frac{\pi t}{T_{\text{max}}}\right)\right)
\end{equation}

where:

\begin{quote}
\begin{quote}
\begin{itemize}
    \item \(\eta_t\) is the learning rate at epoch \(t\)
    \item \(T_{\text{max}} = 120\) (total epochs)
    \item \(\eta_{\text{max}} = 1 \times 10^{-4}\) (initial learning rate)
    \item \(\eta_{\text{min}} = 1 \times 10^{-6}\) (minimum learning rate)
\end{itemize}
\end{quote}
\end{quote}

This schedule provides a smooth, continuous decay from \(\eta_{\text{max}}\) to \(\eta_{\text{min}}\) following a cosine curve, avoiding abrupt learning rate changes and promoting stable convergence. The cosine schedule has been shown to delay difficult denoising tasks until after the midpoint of training, leading to enhanced sample quality and faster convergence compared to linear or step decay schedules. \\

\subsection{Mixed Precision Training}

To accelerate training and reduce memory footprint, Automatic Mixed Precision (AMP) is employed using PyTorch's \texttt{torch.cuda.amp} module:\\
\begin{quote}
\begin{quote}
\begin{itemize}
    \item \textbf{Autocast}: Automatically selects FP16 (half precision) or FP32 (single precision) for each operation based on numerical stability requirements
    \item \textbf{GradScaler}: Scales loss values to prevent gradient underflow in FP16 arithmetic, then unscales gradients before optimizer updates \\
\end{itemize}
\end{quote}
\end{quote}

\newpage
The training loop integrates AMP as follows:\\

\begin{algorithm}
\caption{Training loop approach}\label{alg:cap}
\begin{algorithmic}
\STATE \textbf{Initialize} GradScaler \(\mathcal{S}\)
\FOR{each batch \((\mathbf{x}, \boldsymbol{\epsilon})\)}
    \STATE Zero gradients
    \STATE \textbf{with} autocast():
    \STATE \quad Forward pass: \(\boldsymbol{\epsilon}_\theta \leftarrow \text{model}(\mathbf{x}_t, t/T)\)
    \STATE \quad Compute loss: \(\mathcal{L} = \|\boldsymbol{\epsilon}_\theta - \boldsymbol{\epsilon}\|_2^2\)
    \STATE Scaled backward: \(\mathcal{S}.\text{scale}(\mathcal{L}).\text{backward}()\)
    \STATE Optimizer step: \(\mathcal{S}.\text{step}(\text{optimizer})\)
    \STATE Update scaler: \(\mathcal{S}.\text{update}()\)
    \STATE Update EMA model
\ENDFOR
\end{algorithmic}
\end{algorithm}

Mixed precision training typically yields $1.5$--$3\times$ speedup on modern GPUs (e.g., NVIDIA V100, A100) with Tensor Cores, while maintaining numerical accuracy. This part is covered in more detail in Section 13.2.

\section{Loss Function}

The training objective is the noise prediction mean squared error (MSE):

\begin{equation}
\mathcal{L}_{\text{simple}}(\theta) = \mathbb{E}_{t, \mathbf{x}_0, \boldsymbol{\epsilon}, \mathbf{c}} \left[ \|\boldsymbol{\epsilon} - \boldsymbol{\epsilon}_\theta(\mathbf{x}_t, t, \mathbf{c})\|_2^2 \right]
\label{eq:loss_conditional}
\end{equation}

where:
\begin{quote}
\begin{itemize}
    \item \(t \sim \text{Uniform}\{1, 2, \ldots, T\}\) (random timestep)
    \item \(\mathbf{x}_0 \sim q(\mathbf{x}_0)\) (clean data sample)
    \item \(\mathbf{c}\) (conditioning vector, e.g., wind speed or class label)
    \item \(\boldsymbol{\epsilon} \sim \mathcal{N}(\mathbf{0}, \mathbf{I})\) (Gaussian noise)
    \item \(\mathbf{x}_t = \sqrt{\bar{\alpha}_t} \mathbf{x}_0 + \sqrt{1 - \bar{\alpha}_t} \boldsymbol{\epsilon}\) (noisy sample) \\
\end{itemize}
\end{quote}

This simplified objective (omitting weighting factors) is empirically shown to produce high-quality samples and stable training dynamics. The model learns to denoise by predicting the noise component \(\boldsymbol{\epsilon}\) that was added to the clean data.

\subsection{Context Conditioning}

For the context-conditioned variant, we implement classifier-free guidance training by randomly masking the context during training:

\begin{equation}
c_{\text{masked}} = c \cdot m, \quad m \sim \text{Bernoulli}(0.9)
\end{equation}

where $c$ is the one-hot encoded context vector and $m$ is a binary mask. Each training sample has a 10\% probability of having its context completely zeroed out, allowing the model to learn both conditional and unconditional generation. This enables classifier-free guidance during inference by interpolating between conditional and unconditional predictions.

\section{Data Configuration and Noise Generation}

\subsection{Dataset Characteristics and Preprocessing}
The training dataset consists of single-channel wind field data representing intense convective bursts within rare tropical cyclones. Each data sample $x_{0}$ is a $16 \times 16$ spatial grid, a resolution selected to balance physical detail with computational efficiency for the Context-UNet architecture. The raw imagery is sourced from NASA Global Precipitation Measurement (GPM) and Geostationary Operational Environmental Satellite (GOES) \cite{GOESData} datasets as shown in Figure \ref{fig:data_samples}. \\

The training data consists of:
\begin{quote}
\begin{quote}
\begin{itemize}
    \item \textbf{Source}: Wind field data stored in \texttt{./data/wind\_1D16X16.npy}
    \item \textbf{Data type}: NumPy array, single-channel (1D physical field)
    \item \textbf{Spatial resolution}: \(16 \times 16\) grid points
    \item \textbf{Normalization}: Data values are clipped to \([0, 1]\) range during sample generation
\end{itemize}
\end{quote}
\end{quote}

\subsection{Train-Validation Split}

The dataset is partitioned into training and validation subsets:
\begin{quote}
\begin{quote}
\begin{itemize}
    \item \textbf{Training set}: 90\% of total data
    \item \textbf{Validation set}: 10\% of total data
    \item Splitting performed via \texttt{torch.utils.data.random\_split}
    \item Ensures statistical representativeness with sufficient validation samples
\end{itemize}
latex\begin{figure}[H]
    \centering
    \includegraphics[width=0.80\textwidth]{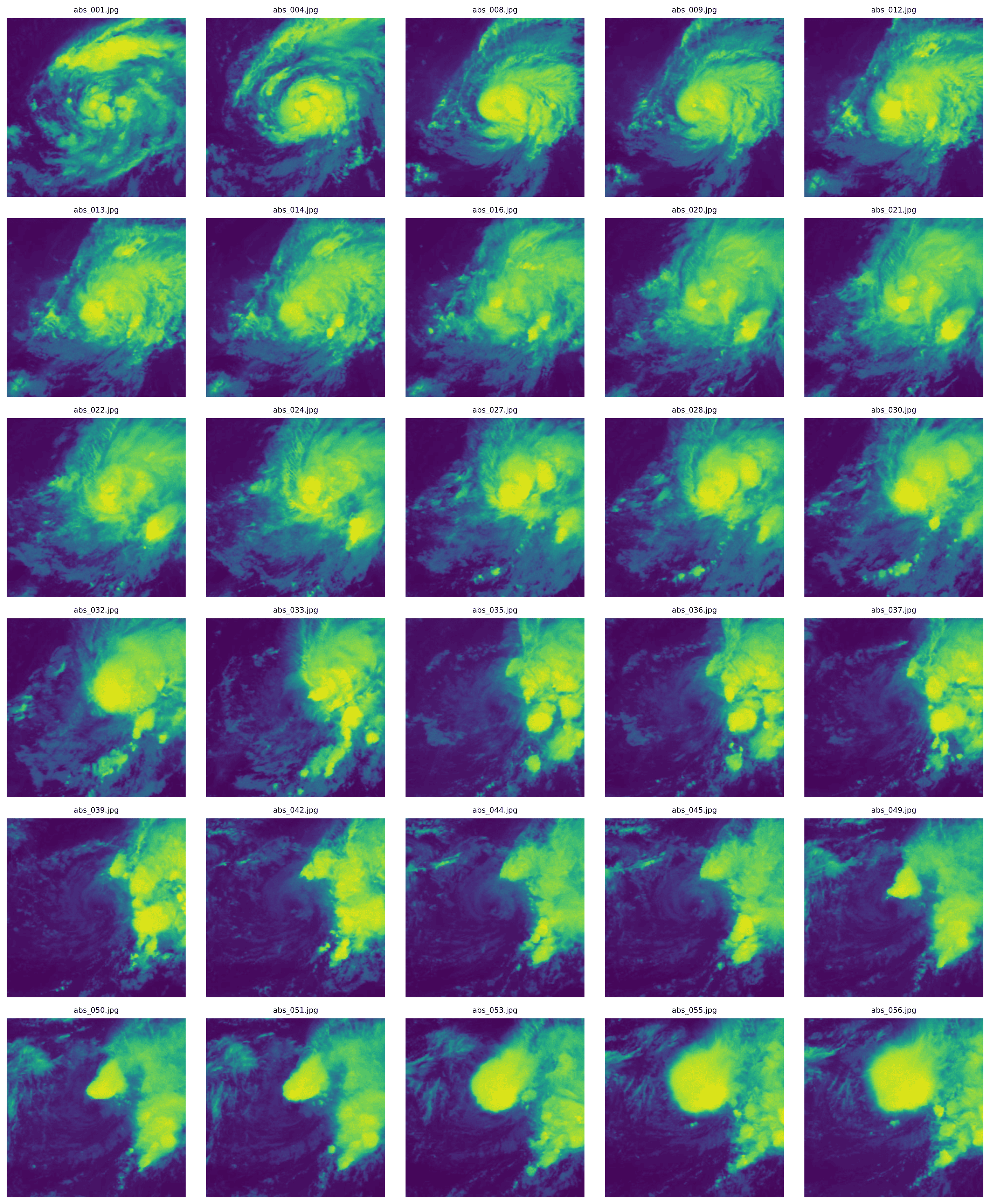} 
    \caption{Cyclone event capture from emergence to dissipation. \textbf{Note:} 
    Images were originally $1024 \times 1024$ pixels and have been cropped to 
    $16 \times 16$ pixels for training.}
    \label{fig:data_samples}
\end{figure}

\end{quote}
\end{quote}

\begin{figure}[H]
    \centering
    \includegraphics[width=0.8\linewidth]{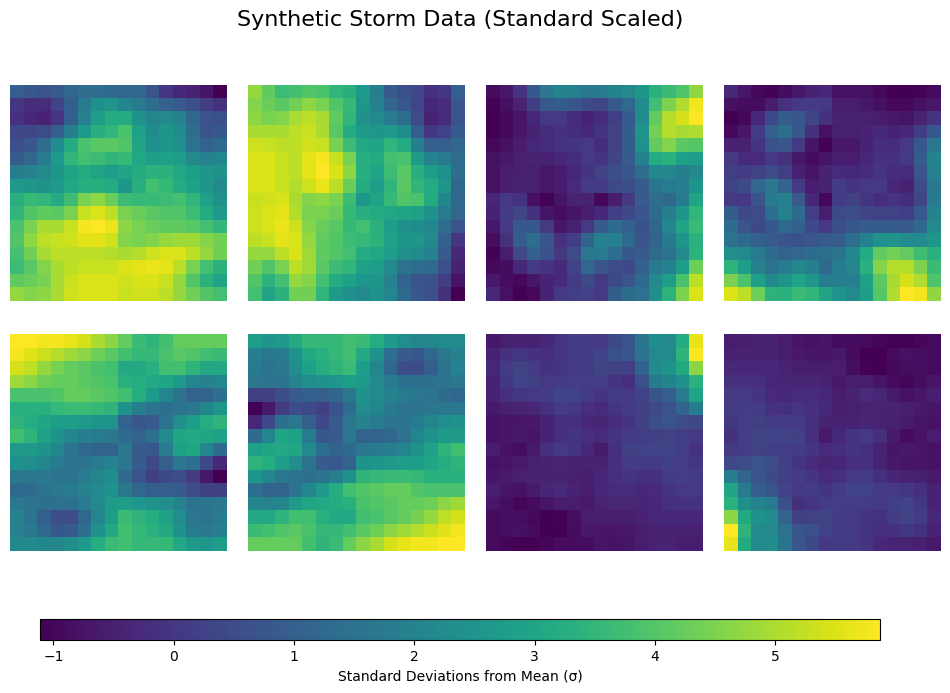}
    \caption{Storm Wind Fields representing physical intensity on a $16 \times 16$ grid. Samples are conditioned on specific atmospheric parameters (Param 0 and 1) representing Rapid Intensification (RI) conditions. The colormap highlights the high-intensity gradients preserved by the physics-informed model.}
    \label{fig:storm_samples}
\end{figure}

To ensure numerical stability during the diffusion process, all input values are normalized using a \textit{StandardScaler} to remove mean and unit variance bias before being clipped to a $[0, 1]$ range for consistency. As shown in Figure \ref{fig:storm_samples}, the resulting intensity fields exhibit high-gradient spatial features characteristic of rapid intensification (RI).

\subsection{Label Distribution and Data Scarcity}
A significant challenge addressed in this work is the scarcity of data for extreme rare weather events. To categorize the training data, we employ a multi-class labeling system based on physical parameters defining RI, such as low vertical wind shear and high ocean heat content. 
\begin{figure}[H]
    \centering
    \includegraphics[width=0.85\linewidth]{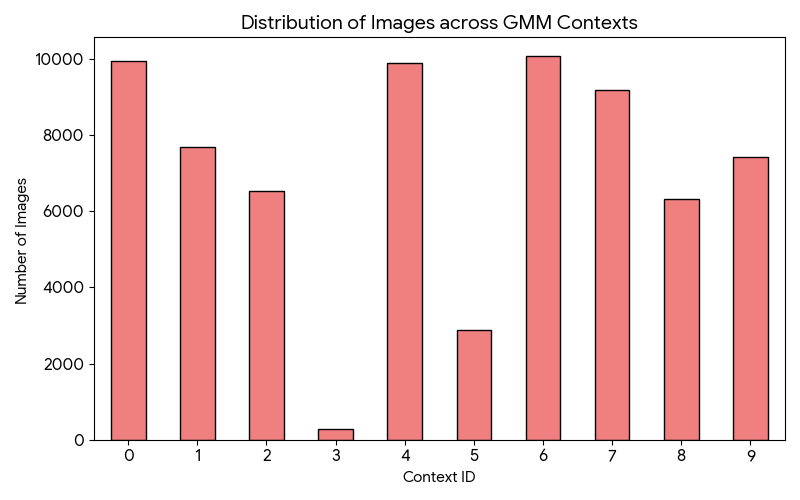}
    \caption{Distribution of unique physical parameter labels within the dataset. The extreme scarcity of samples in Class 4 (202 samples) demonstrates the data bottleneck for rare weather events that this research aims to mitigate through synthetic augmentation.}
    \label{fig:label_dist}
\end{figure}
In the context of training diffusion models with grayscale images, the choice of scaling is critical because diffusion processes---specifically the forward and reverse Gaussian noise additions---are mathematically designed to operate on a specific range. While basic image processing often uses Min-Max to scale pixels to $[0, 1]$, diffusion models almost exclusively use Min-Max to scale pixels to the $[-1, 1]$ range.\\

For almost all modern diffusion architectures (like DDPM or Stable Diffusion) \cite{rombach_high-resolution_2022}, Min-Max Scaling is the standard approach, but with a specific target range.
DDPM (Denoising Diffusion Probabilistic Models) paper by Ho et al. (2020)
Figure \ref{fig:label_dist} above illustrates the distribution of these unique physical parameter labels across the training set. The distribution reveals a stark imbalance, with extreme classes (e.g., Category 4 and 5 equivalents) significantly underrepresented compared to baseline storm conditions. Specifically, Class 0 contains approximately 79,768 samples, while the most extreme rare event class (Class 4) contains only 202 samples. This 400-fold difference in class frequency underscores the critical need for the proposed generative model to augment training data for downstream operational detection algorithms.

\section{Model Architecture}

\subsection{Network Configuration}

The model employs a Context U-Net architecture designed for conditional diffusion modeling. The configuration parameters are presented in Table \ref{tab:model_config}.

\begin{table}[h]
\centering
\begin{tabular}{ll}
\toprule
\textbf{Parameter} & \textbf{Value} \\
\midrule
Architecture & Context U-Net \\
Input channels & 1 (grayscale) \\
Feature dimension ($n_{\text{feat}}$) & 64 \\
Context features ($n_{\text{cfeat}}$) & 5 (standard), 10 (context) \\
Image resolution & 16 $\times$ 16 pixels \\
Activation function & SiLU (Swish) \\
Normalization & Group Normalization \\
\bottomrule
\end{tabular}
\caption{Model architecture configuration parameters.}
\label{tab:model_config}
\end{table}

\subsection{Exponential Moving Average}

To improve sample quality and training stability, we maintain an exponential moving average (EMA) of the model weights:

\begin{equation}
\theta_{\text{EMA}}^{(t)} = \beta_{\text{EMA}} \cdot \theta_{\text{EMA}}^{(t-1)} + (1 - \beta_{\text{EMA}}) \cdot \theta^{(t)}
\end{equation}

where $\beta_{\text{EMA}} = 0.995$ is the decay factor, $\theta^{(t)}$ are the model parameters at step $t$, and $\theta_{\text{EMA}}^{(t)}$ are the EMA parameters. The EMA model is updated after each gradient step and is used for inference and sample generation, while the standard model is used for training.

\section{Training Methodology}

\subsection{Training Hyperparameters}

The complete set of training hyperparameters is detailed in Table \ref{tab:training_params}.

\begin{table}[h]
\centering
\begin{tabular}{ll}
\toprule
\textbf{Parameter} & \textbf{Value} \\
\midrule
Batch size & 64 \\
Training epochs & 120 \\
Initial learning rate & $10^{-4}$ \\
Optimizer & Adam \\
Adam $\beta_1$ & 0.9 \\
Adam $\beta_2$ & 0.999 \\
Weight decay & 0 \\
Learning rate scheduler & Cosine Annealing \\
Minimum learning rate & $10^{-6}$ \\
EMA decay rate & 0.995 \\
Data split ratio & 90\% train, 10\% validation \\
Context masking probability & 0.1 (context model) \\
\bottomrule
\end{tabular}
\caption{Training hyperparameters and optimization settings.}
\label{tab:training_params}
\end{table}

\subsection{Learning Rate Schedule}

A cosine annealing learning rate schedule is employed to gradually reduce the learning rate over training:

\begin{equation}
\eta_t = \eta_{\min} + \frac{1}{2}(\eta_{\max} - \eta_{\min})\left(1 + \cos\left(\frac{t}{T_{\max}}\pi\right)\right)
\end{equation}

where $\eta_{\max} = 10^{-4}$ is the initial learning rate, $\eta_{\min} = 10^{-6}$ is the minimum learning rate, $T_{\max} = 120$ represents the total number of epochs, and $t$ is the current epoch. This schedule provides smooth decay from the initial learning rate to the minimum, promoting stable convergence in later training stages.

\section{Training Procedure}

\subsection{Training Loop (Per Epoch)}

\begin{algorithm}
\caption{Training Loop for Diffusion Model}

\begin{algorithmic}[1]
\STATE Initialize running loss \(\mathcal{L}_{\text{train}} = 0\)
\FOR{each batch \((\mathbf{x}, \boldsymbol{\epsilon})\) in training DataLoader}
    \STATE Zero optimizer gradients
    \STATE Move data to GPU: \(\mathbf{x} \leftarrow \mathbf{x}.\text{to}(\text{device})\), \(\boldsymbol{\epsilon} \leftarrow \boldsymbol{\epsilon}.\text{to}(\text{device})\)
    \STATE Sample random timesteps: \(t \sim \text{Uniform}\{1, \ldots, 500\}\) for each batch element
    \STATE Perturb input: \(\mathbf{x}_t \leftarrow \texttt{perturb\_input}(\mathbf{x}, t, \boldsymbol{\epsilon})\)
    \STATE \textbf{with} autocast():
    \STATE \quad Predict noise: \(\boldsymbol{\epsilon}_\theta \leftarrow \text{model}(\mathbf{x}_t, t / 500)\)
    \STATE \quad Compute loss: \(\mathcal{L} \leftarrow \text{MSE}(\boldsymbol{\epsilon}_\theta, \boldsymbol{\epsilon})\)
    \STATE Scaled backward pass: \(\texttt{scaler.scale}(\mathcal{L}).\text{backward}()\)
    \STATE Optimizer step: \(\texttt{scaler.step}(\text{optimizer})\)
    \STATE Update scaler: \(\texttt{scaler.update}()\)
    \STATE Update EMA model: \(\theta_{\text{EMA}} \leftarrow 0.995 \cdot \theta_{\text{EMA}} + 0.005 \cdot \theta\)
    \STATE Accumulate loss: \(\mathcal{L}_{\text{train}} \leftarrow \mathcal{L}_{\text{train}} + \mathcal{L}\)
\ENDFOR
\STATE Compute average training loss: \(\bar{\mathcal{L}}_{\text{train}} = \mathcal{L}_{\text{train}} / N_{\text{batches}}\)
\STATE Step learning rate scheduler: \(\texttt{scheduler.step}()\)
\end{algorithmic}
\end{algorithm}

\newpage

\section{Sampling and Inference}

\subsection{Generation Procedure}

Sample generation employs the EMA model and follows the reverse diffusion process:

\subsection{Validation Loop (Per Epoch)}

\begin{algorithm}
\caption{Validation Step for Diffusion Model}
\begin{algorithmic}[1]
\STATE Set model to evaluation mode: \texttt{model.eval()}
\STATE Initialize validation loss \(\mathcal{L}_{\text{val}} = 0\)
\STATE \textbf{with} \texttt{torch.no\_grad()}:
\FOR{each batch \(\mathbf{x}\) in validation DataLoader}
    \STATE Move data to GPU and cast to float32
    \STATE Extract single channel: \(\mathbf{x} \leftarrow \mathbf{x}[:, 0:1, :, :]\)
    \STATE Sample random timesteps: \(t \sim \text{Uniform}\{1, \ldots, 500\}\)
    \STATE Sample noise: \(\boldsymbol{\epsilon} \sim \mathcal{N}(\mathbf{0}, \mathbf{I})\)
    \STATE Perturb input: \(\mathbf{x}_t \leftarrow \texttt{perturb\_input}(\mathbf{x}, t, \boldsymbol{\epsilon})\)
    \STATE Predict noise: \(\boldsymbol{\epsilon}_\theta \leftarrow \texttt{model}(\mathbf{x}_t, t/500)\)
    \STATE Compute loss: \(\mathcal{L} \leftarrow \text{MSE}(\boldsymbol{\epsilon}_\theta, \boldsymbol{\epsilon})\)
    \STATE Accumulate loss: \(\mathcal{L}_{\text{val}} \leftarrow \mathcal{L}_{\text{val}} + \mathcal{L}\)
\ENDFOR
\STATE Compute average validation loss: \(\bar{\mathcal{L}}_{\text{val}} = \mathcal{L}_{\text{val}} / N_{\text{batches}}\)
\STATE Set model back to training mode: \texttt{model.train()}
\end{algorithmic}
\end{algorithm}

\textbf{Note}: Validation evaluates the standard model \(\theta\), not the EMA model \(\theta_{\text{EMA}}\), to monitor the primary training trajectory.

\begin{algorithm}
\caption{Sample Generation for Diffusion Model}
\begin{algorithmic}[1]
\STATE Set EMA model to evaluation mode: \(\text{ema\_model.eval}()\)
\STATE Initialize from pure noise: \(\mathbf{x}_T \sim \mathcal{N}(\mathbf{0}, \mathbf{I})\), shape \((N, 1, 16, 16)\)
\FOR{\(i = T, T-1, \ldots, 1\)}
    \STATE Normalize timestep: \(t \leftarrow i / T\)
    \IF{\(i > 1\)}
        \STATE Sample noise: \(\mathbf{z} \sim \mathcal{N}(\mathbf{0}, \mathbf{I})\)
    \ELSE
        \STATE Set deterministic: \(\mathbf{z} \leftarrow \mathbf{0}\)
    \ENDIF
    \STATE Predict noise: \(\boldsymbol{\epsilon}_\theta \leftarrow \text{ema\_model}(\mathbf{x}_i, t)\)
    \STATE Denoise: \(\mathbf{x}_{i-1} \leftarrow \texttt{denoise\_add\_noise}(\mathbf{x}_i, i, \boldsymbol{\epsilon}_\theta, \mathbf{z})\)
\ENDFOR
\STATE Clamp to valid range: \(\mathbf{x}_0 \leftarrow \text{clamp}(\mathbf{x}_0, 0, 1)\)
\STATE Save as image grid: \(\texttt{save\_image}(\mathbf{x}_0, \text{path}, \text{nrow}=4)\)
\end{algorithmic}
\end{algorithm}

This procedure generates \(N=16\) samples by default, arranged in a \(4 \times 4\) grid. Samples are generated every 4 epochs during training for qualitative assessment.


\begin{table}[h]
\centering
\caption{Complete Training Configuration Summary}
\begin{tabular}{@{}ll@{}}
\toprule
\textbf{Component} & \textbf{Specification} \\ \midrule
\multicolumn{2}{c}{\textit{Model Architecture}} \\
Model & Context U-Net \\
Input channels & 1 (grayscale) \\
Base features & 64 \\
Context features & 5 \\
Spatial resolution & $16 \times 16$ \\
\midrule
\multicolumn{2}{c}{\textit{Training Configuration}} \\
Batch size & 64 \\
Epochs & 120 \\
Optimizer & Adam (\(\beta_1=0.9\), \(\beta_2=0.999\)) \\
Learning rate (initial) & \(1 \times 10^{-4}\) \\
Learning rate (minimum) & \(1 \times 10^{-6}\) \\
LR scheduler & Cosine Annealing \\
EMA decay & 0.995 \\
Loss function & MSE (noise prediction) \\
Mixed precision & Enabled (autocast + GradScaler) \\
\midrule
\multicolumn{2}{c}{\textit{Checkpointing}} \\
Frequency & Every 4 epochs + final \\
Saved models & Standard + EMA \\
Generated samples & 16 ($4 \times 4$ grid) \\

\bottomrule
\end{tabular}
\end{table}

\newpage

\section{Results}
\label{sec:results}

To evaluate the model's ability to generate class-specific wind patterns, we conducted conditional generation experiments using one-hot encoded context vectors. The model was trained with ncfeat = 10 context features, corresponding to 10 distinct wind pattern classes (labeled 0--9). The model architecture consists of a U-Net backbone with contextual embedding layers, trained for 500 diffusion timesteps using a linear beta schedule ranging from $\beta_1 = 10^{-4}$ to $\beta_2 = 0.02$. All experiments were conducted on $16 \times 16$ grayscale wind field images, with pixel values normalized to the range $[-1, 1]$ using a min-max scaler fitted on the training dataset. The dataset consists of different storm characteristics organized into two ocean groups.

\subsection{Context-Conditioned Generation}
Ocean Group A encompasses storms primarily forming in Ocean 1, following a clear lifecycle progression from early development through peak intensity to eventual dissipation. The journey begins with nascent systems (Contexts 7 \& 9) averaging 33 hours old with winds around 47 knots, then strengthens through Context 0 at approximately 50 hours and 51 knots. These storms reach their peak maturity in Context 1 around the 5-day mark (120 hours), achieving maximum average winds of 56 knots for this ocean group. As systems age beyond this point, they enter prolonged weakening phases (Contexts 6 \& 5) spanning 6--10 days with diminishing winds of 41--43 knots. Notably, Context 3 represents an exceptional subset of extraordinarily persistent storms that survive beyond 17 days (416 hours), though these rare events constitute only a small fraction of Ocean 1's storm population.\\

Ocean Group B, dominated by Ocean 2 systems, exhibits a more concentrated development pattern with notably higher intensities at maturity. The majority of these storms (Contexts 2 \& 4) are captured during their formative stages around 41 hours old, displaying wind speeds between 47--53 knots---comparable to early Ocean 1 development but progressing through a different evolutionary trajectory. The defining characteristic of Ocean Group B emerges in Context 8, which represents large, exceptionally powerful mature cyclones that maintain their strength far longer than their Ocean 1 counterparts. These Context 8 systems reach an impressive average of 65 knots---the highest wind speed across all contexts---while persisting for approximately one week (168 hours), demonstrating Ocean 2's capacity to sustain significantly more intense storms at extended durations compared to Ocean Group A's typical lifecycle.\\

\begin{table}[h]
\centering
\caption{Storm lifecycle characteristics across different contexts and ocean basins}
\label{tab:storm_contexts}
\begin{tabular}{@{}ccccp{6cm}@{}}
\toprule
\textbf{Context} & \textbf{Ocean} & \textbf{Stage} & \textbf{Avg. Wind Speed} & \textbf{Description} \\
\midrule
7, 9 & 1 & Early & 47 kn & Initial formation/Early development \\
0 & 1 & Mid-Early & 51 kn & Intensifying tropical system \\
1 & 1 & Mature & 56 kn & Peak intensity phase \\
6, 5 & 1 & Late & 42 kn & Long-duration, weakening storms \\
3 & 1 & Extreme & 43 kn & Rare, ultra-long-lived systems \\
\midrule
2, 4 & 2 & Early & 50 kn & Standard development in Ocean 2 \\
8 & 2 & Late-Peak & 65 kn & Major mature storms (Typhoons/Cyclones) \\
\bottomrule
\end{tabular}
\end{table}

To evaluate the model's ability to generate class-specific wind patterns, we conducted conditional generation experiments using one-hot encoded context vectors. The model was trained with $n_{\text{cfeat}} = 10$ context features, corresponding to 10 distinct wind pattern classes (labeled 0--9). \\

\begin{figure}[htbp]
    \centering
    \includegraphics[width=1.0\textwidth]{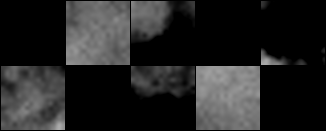}
    \caption{Context-conditioned generation for labels 0-9. Each image represents a wind field pattern generated conditioned on its corresponding class label using one-hot encoding. The model successfully produces distinct patterns for different context inputs.}
    \label{fig:standard_model}
\end{figure}

Figure~\ref{fig:standard_model} demonstrates the model's capacity for controlled generation across all 10 context classes. Each sample corresponds to a specific context label, generated by passing the respective one-hot encoded vector through the context conditioning mechanism.

The results in Figure~\ref{fig:standard_model} reveal that the model learns discriminative features for each context class. Classes 2,4 and 7 produce more structured patterns with pronounced intensity variations. This demonstrates the effectiveness of the context embedding architecture in capturing class-specific characteristics during the denoising process.

 \newpage

\begin{figure}[htbp]
    \centering
    \subfloat[Samples for context label 1]{
        \includegraphics[width=0.48\textwidth]{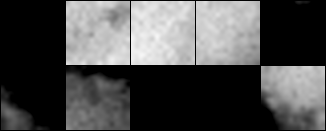}
        \label{fig:context1}
    }
    \hfill
    \subfloat[Samples for context label 5]{
        \includegraphics[width=0.48\textwidth]{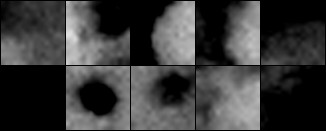}
        \label{fig:context5}
    }
    \caption{Multiple samples generated for specific context labels. (a) Context label 1 produces predominantly low-intensity, smooth gradient fields. (b) Context label 5 generates high-contrast patterns with distinct spatial structures and intensity clusters.}
    \label{fig:context_specific}
\end{figure}

To further investigate the model's learned representations, we analyzed the generation quality for individual context classes. Figures~\ref{fig:context1} and~\ref{fig:context5} show multiple samples generated for context labels 1 and 5, respectively.\\

Qualitative assessment of the generated outputs reveals a significant correlation between the conditioning level and the structural complexity of the wind fields. At a lower conditioning level (Context 1), the model produces relatively uniform, high-frequency patterns characterized by soft morphology and low contrast, suggesting a steady-state or calm atmospheric representation. In contrast, Context 5 triggers the generation of highly structured features, including distinct localized "cells" and vortex-like structures with sharp gradients. These results indicate that the contextual embedding layers effectively modulate the U-Net's feature maps, enabling the model to transition from generating basic Gaussian-like noise to complex, turbulent-like meteorological states as the context value increases.\\

\begin{figure}[htbp]
    \centering
    \subfloat[Samples for context label 8 at epoch 116]{
        \includegraphics[width=0.48\textwidth]{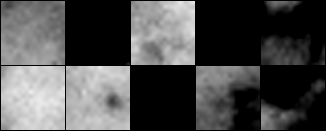}
        \label{fig:context8_epoch116}
    }
    \hfill
    \subfloat[Samples for context label 8 at epoch 4]{
        \includegraphics[width=0.48\textwidth]{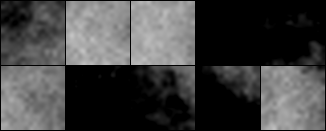}
        \label{fig:context8_epoch4}
    }
    
    \caption{As the conditioning increases to epoch 116, the U-Net backbone successfully synthesizes well-defined localized gradients and cellular structures. At lower conditioning levels, the model produces high-frequency, low-contrast outputs indicative of early-stage training or weak prior guidance.}
    \label{fig:context8_comparison}
\end{figure}

A comparison of the generated wind fields across various conditioning states reveals a clear progression in structural resolution and morphological detail. Figures~\ref{fig:context8_epoch116} and~\ref{fig:context8_epoch4} compare samples for Context 8 at epochs 116 and 4, respectively.\\

At lower training checkpoints (epoch 4) or lower-order conditioning (Context 8), the model exhibits limited capacity for high-fidelity reconstruction, resulting in "blurry" or "hazy" outputs characterized by high-frequency noise and poorly defined boundaries. These early-stage results represent a global mean of the training data, where the U-Net backbone has yet to learn the localized, sharp gradients necessary for realistic wind modeling.\\

As conditioning increases through training (epoch 116), the model successfully transitions from stochastic noise to coherent structural features. This is evidenced by the emergence of distinct localized "eyes" and cellular vortex structures, suggesting that the contextual embedding layers effectively guide the denoising process to capture the non-linear fluid dynamics inherent in the storm dataset. Despite the coarse $16 \times 16$ pixel resolution, the higher-context samples demonstrate a marked improvement in contrast and spatial organization, effectively utilizing the limited pixel space to represent complex atmospheric phenomena.\\

\newpage

Although visual inspection demonstrates qualitative success, we note several key observations regarding model performance.

\begin{quote}
\begin{quote}
\begin{itemize}
    \item \textbf{Spatial coherence}: Generated samples maintain realistic spatial autocorrelation patterns typical of wind field data, avoiding checkerboard artifacts or high-frequency noise.
    
    \item \textbf{Diversity}: The model produces varied samples within each context class, suggesting adequate mode coverage and avoiding mode collapse.
    
    \item \textbf{Context fidelity}: The clear visual distinctions between context classes indicate successful conditioning, with the model responding appropriately to different one-hot encoded inputs.
    
    \item \textbf{Scaling consistency}: The inverse StandardScaler transformation successfully maps generated samples back to the physically meaningful [0, 255] intensity range, with proper clipping to handle edge cases.
\end{itemize}
\end{quote}
\end{quote}
The results demonstrate that the proposed context-conditioned DDPM architecture effectively learns to generate realistic wind field patterns while maintaining controllability through context conditioning. 

\subsection{Implementation Details}
\label{sec:implementation}

All experiments were implemented in PyTorch 2.0+ and executed on NVIDIA GPUs with CUDA support. The denoising model architecture consists of a ContextUnet with 64 base features ($n_{\text{feat}} = 64$), processing single-channel grayscale images of size $16 \times 16$ pixels. The diffusion process employs 500 timesteps with a linear variance schedule, and generation is performed using the DDPM sampling algorithm without classifier-free guidance.\\ 

The model achieved a Log-Spectral Distance (LSD) of 4.5 dB, indicating that the generated samples successfully captured the fundamental global structure \cite{oord_wavenet_2016}. Although this score reflects a strong statistical alignment, the remaining distance suggests minor discrepancies in high-frequency texture or fine-grained detail compared to the ground truth. Data pre-processing involves flattening each $16 \times 16$ image to a 256-dimensional vector, followed by standardization using scikit-learn's StandardScaler. The scaler parameters (mean and standard deviation) are fitted on the training set and saved for consistent inverse transformation during inference. Context conditioning is achieved through one-hot encoding of class labels, which are embedded and concatenated with intermediate U-Net features at multiple resolution levels.\\

The sampling procedure initializes with Gaussian noise $\mathbf{x}_T \sim \mathcal{N}(\mathbf{0}, \mathbf{I})$ and iteratively denoises according to:
\begin{equation}
\mathbf{x}_{t-1} = \frac{1}{\sqrt{\alpha_t}} \left( \mathbf{x}_t - \frac{1-\alpha_t}{\sqrt{1-\bar{\alpha}_t}} \boldsymbol{\epsilon}_\theta(\mathbf{x}_t, t, \mathbf{c}) \right) + \sqrt{\beta_t} \mathbf{z}, \quad \mathbf{z} \sim \mathcal{N}(\mathbf{0}, \mathbf{I})
\end{equation}
where $\boldsymbol{\epsilon}_\theta$ is the learned noise predictor, $\mathbf{c}$ is the context vector, $\alpha_t = 1 - \beta_t$, and $\bar{\alpha}_t = \prod_{s=1}^t \alpha_s$. The code is made available at \texttt{[https://github.com/MarawanYakout/SERWED.git]} for reproducibility.

\section{Discussion}

Our context-conditioned diffusion model effectively addresses the extreme class imbalance in meteorological datasets. The nearly 400-fold disparity between Class 0 (79,768 samples) and Class 4 (202 samples) presents a fundamental challenge for supervised learning. Traditional augmentation techniques (rotation, flipping, brightness adjustment) merely create variations of existing samples and can violate physical constraints inherent in atmospheric data.\\

By learning the underlying generative process, our model generates genuinely novel samples that maintain physical and statistical characteristics of each storm class. This approach provides particular value for rare extreme events where limited examples make traditional augmentation insufficient. 
Compared to standard GAN-based augmentation, our physics-informed diffusion approach avoids mode collapse and provides higher sample diversity, which is crucial for training downstream detection models that must generalize across varying storm morphologies.\\

A critical requirement for synthetic meteorological data is adherence to physical constraints. The context-conditioning mechanism embeds key atmospheric parameters, including wind shear, ocean heat content, and storm age, directly into the generation process, ensuring that synthesized samples reflect realistic relationships between environmental conditions and storm characteristics. Qualitative analysis reveals clear differentiation across context classes. Low-intensity contexts (0, 1, 7, 9) generate smooth gradient fields characteristic of developing or weakening systems, while high-intensity contexts (4, 8) produce localized high-gradient regions and pronounced vortex structures. The progression from Context 0 (51 knots) through Context 1 (56 knots) to Context 8 (65 knots) demonstrates that the model captures realistic storm intensification patterns rather than overfitting to specific training examples. \\ 

Additionally, the coherent spatial organization of generated samples, combined with the absence of checkerboard artifacts or nonphysical high-frequency noise, suggests that the model has learned spatially correlated wind field structures rather than arbitrary pixel-level patterns. This is particularly important for applications requiring physically interpretable synthetic data.\\

The progression from epoch 4 to epoch 116 illustrates the model's evolving capacity to capture fine-grained atmospheric structures. Early in training (epoch 4), generated samples exhibit high-frequency noise and lack coherent spatial organization, characteristic of under-fitted generative models. By epoch 116, the model generates well-defined vortex structures and sharp intensity gradients, indicating successful learning of hierarchical atmospheric features. This evolution demonstrates that the Context-UNet architecture effectively learns multi-scale representations. Skip connections preserve spatial coherence while the encoder-decoder pathway captures increasingly abstract structural patterns. These observations align with theoretical expectations for diffusion models, where iterative denoising enables progressive refinement of spatial structure.\\

Several methodological design choices were influenced by the need to balance computational feasibility with model performance:\\

\begin{quote}
\begin{itemize}

 \item \textbf{Pre-generated Noise Strategy}- Ensures fair representation of rare classes by providing consistent training conditions across all 120 epochs. Each image faces identical denoising challenges, eliminating variance that could disproportionately affect learning for the 202 Class 4 examples. The trade-off is substantial storage requirements and reduced stochasticity in training.\\

 \item \textbf{Spatial resolution}- The $16 \times 16$ resolution balances computational feasibility with physical detail. While this sacrifices fine-scale features from the original $1024 \times 1024$ imagery, it enabled efficient training and rapid iteration. Higher resolutions would quadratically increase memory requirements.\\

 \item \textbf{Mixed-precision training}- Achieved $1.5$--$3\times$ speedup on modern GPUs with Tensor Cores while maintaining numerical stability, partially addressing the high computational costs inherent in diffusion model training.

\end{itemize}
\end{quote}

\subsection{Computational Limitations}
Training diffusion models presents significant computational challenges, as noted in our introduction. The high computational cost stems from several factors. The model requires 120 epochs over 140,514 samples, with each sample processed at multiple random timesteps (1-500), necessitating repeated forward and backward passes through the U-Net for computing noise schedules and gradients.\\

The pregenerated noise strategy, while ensuring reproducibility and fair representation of rare classes, requires storing a large tensor (140,514 images $\times$ 500 timesteps $\times$ 1 channel $\times$ 16 $\times$ 16 pixels), consuming substantial disk space. The $16 \times 16$ resolution was chosen specifically to make training computationally feasible; scaling to higher resolutions would dramatically increase requirements---moving to $64 \times 64$ would increase memory by $16\times$ and training time by approximately $10-15\times$. While training is computationally intensive, generation is more efficient once the model is trained. However, each generated sample still requires 500 forward passes through the U-Net (one per denoising step), precluding real-time generation for operational deployment. \\

These computational constraints influenced our architectural choices and represent practical limitations for scaling this approach to operational forecasting systems that require higher-resolution images or real-time generation capabilities. Beyond computational constraints, three additional limitations require acknowledgment. The spatial resolution of $16 \times 16$ sacrifices fine-scale atmospheric features such as mesoscale convection and detailed eye structure. Moreover, single-timestep generation cannot capture temporal storm evolution.\\ 

Finally, while conditioning guides generation toward physical plausibility, physical relationships are learned implicitly from data rather than explicitly enforced.
To remain within the T4's 16GB VRAM envelope, we purposefully balanced the Context-UNet's depth ($n_{feat}=64$) with a compressed $16\times16$ spatial manifold. This allowed for a 500-step diffusion trajectory that prioritizes thermodynamic consistency over raw pixel density.

\subsection{Future Directions}
Several directions emerge as important avenues for future investigation. First, scaling to higher resolutions ($64 \times 64$ or $128 \times 128$) would capture the mesoscale features currently lost at 16×16 resolution, potentially using progressive training strategies to manage the associated computational costs. Second, extending the model to time-series generation would enable modeling of storm evolution and intensification dynamics over multiple timesteps rather than single snapshots. Third, incorporating physics-informed loss functions that explicitly enforce physical constraints such as mass continuity and momentum balance would provide stronger consistency guarantees beyond the implicit learning achieved through data-driven conditioning.\\

This methodology applies to other rare atmospheric phenomena (tornadogenesis, flash flooding, severe convection) and, more broadly, to any domain where extreme class imbalance limits ML development for applications governed by physical laws. By combining generative modeling with domain-specific conditioning, we can expand limited datasets while maintaining physical fidelity.

\section{Conclusions}

We have developed and validated a physics-informed, context-conditioned diffusion model for generating synthetic wind field data associated with rare tropical cyclone events. By integrating key atmospheric parameters directly into the Context-UNet conditioning mechanism, we ensure that generated samples reflect realistic relationships between environmental conditions and storm characteristics. Our methodology, which includes a pre-generated noise strategy, mixed-precision training, and cosine annealing learning rate schedule, achieved stable convergence and produced spatially coherent outputs.\\

A central contribution of this work is to address the severe class imbalance inherent in rapid intensification datasets, where rare extreme events are vastly underrepresented. By learning the underlying generative process rather than relying on conventional augmentation, the model is able to synthesize novel, physically consistent examples of rare events, thereby mitigating a critical bottleneck in extreme weather machine learning applications.\\

The results further indicate that the model successfully learns discriminative features across ten distinct context classes, with clear differentiation between low-intensity developing storms and high-intensity mature cyclones. The training progression from epoch 4 to epoch 116 shows the model's evolution from producing high-frequency noise to generating well-defined vortex structures with sharp intensity gradients characteristic of rapidly intensifying systems.\\

Future work should focus on scaling to higher spatial resolutions to capture mesoscale atmospheric features, extending to temporal sequences to model storm evolution over time, and incorporating physics-informed constraints. The methodology presented here is broadly applicable to other rare atmospheric phenomena and, more generally, to any domain where extreme class imbalance limits machine learning model development for applications governed by known physical laws.\\

\newpage
\begin{adjustwidth}{-\extralength}{0cm}

\reftitle{References}

\end{adjustwidth}
\end{document}